\documentclass[runningheads]{llncs}
\usepackage{graphicx}
\usepackage{multirow}
\usepackage[misc]{ifsym}
\usepackage{hyperref}

\makeatletter
\newcommand{\printfnsymbol}[1]{%
  \textsuperscript{\@fnsymbol{#1}}%
}
\makeatother

\begin{document}
\title{MLDT: Multi-Level Decomposition for Complex Long-Horizon Robotic Task Planning with Open-Source Large Language Model}
\titlerunning{MLDT: Multi-Level Decomposition for Robotic Task Planning}
\author{Yike Wu\inst{1,2}\thanks{Both authors contributed equally to this research.}\
\and Jiatao Zhang\inst{3}\printfnsymbol{1}\
\and Nan Hu\inst{1,2}\
\and Lanling Tang\inst{4}\
\and Guilin Qi\inst{1,2\textsuperscript{(\Letter)}}\
\and Jun Shao\inst{3}\
\and Jie Ren\inst{5}\
\and Wei Song\inst{5\textsuperscript{(\Letter)}}\
}
\authorrunning{Wu and Zhang, et al.}
\institute{Southeast University, Nanjing, Jiangsu, China\\ \email{\{yike.wu,nanhu,gqi\}@seu.edu.cn} \and Key Laboratory of New Generation Artificial Intelligence Technology and Its Interdisciplinary Applications (Southeast University), Ministry of Education \and Zhejiang University, Hangzhou, Zhejiang, China\\ \email{\{jiataozh,jun\_shao\}@zju.edu.cn} \and
University of Chinese Academy of Sciences, Beijing, China\\ \email{tanglanling22@mails.ucas.ac.cn} \and
Zhejiang Lab, Hangzhou, Zhejiang, China\\ \email{\{renjie,weisong\}@zhejianglab.com}}
\maketitle
\setcounter{footnote}{0}
\begin{abstract}
In the realm of data-driven AI technology, the application of open-source large language models (LLMs) in robotic task planning represents a significant milestone. Recent robotic task planning methods based on open-source LLMs typically leverage vast task planning datasets to enhance models' planning abilities. While these methods show promise, they struggle with complex long-horizon tasks, which require comprehending more context and generating longer action sequences. This paper addresses this limitation by proposing MLDT, the \textbf{M}ulti-\textbf{L}evel \textbf{D}ecomposition \textbf{T}ask planning method. This method innovatively decomposes tasks at the goal-level, task-level, and action-level to mitigate the challenge of complex long-horizon tasks. In order to enhance open-source LLMs' planning abilities, we introduce a goal-sensitive corpus generation method to create high-quality training data and conduct instruction tuning on the generated corpus. Since the complexity of the existing datasets is not high enough, we construct a more challenging dataset, LongTasks, to specifically evaluate planning ability on complex long-horizon tasks. We evaluate our method using various LLMs on four datasets in VirtualHome. Our results demonstrate a significant performance enhancement in robotic task planning, showcasing MLDT's effectiveness in overcoming the limitations of existing methods based on open-source LLMs as well as its practicality in complex, real-world scenarios.\footnote{Our code is available at https://github.com/wuyike2000/MLDT}

%Recent robotic task planning methods based on open-source large language models (LLMs) typically incorporate task-related information to generate action sequences grounded in the environment. However, these methods face challenges on complex long-horizon tasks, which require comprehending more context and generating longer action sequences. This paper addresses this limitation by proposing MLDT, \textbf{M}ulti-\textbf{L}evel \textbf{D}ecomposition \textbf{T}ask planning method. This method innovatively decomposes tasks into goal, task, and action levels to mitigate the challenge of complex long-horizon tasks. In order to enhance open-source LLMs' planning ability, we introduce a goal-sensitive corpus generation method to create high-quality training data compatible with our method and conduct instruction tuning on the generated corpus. Since the complexity of the existing datasets is not high enough, we construct a more challenging dataset, LongTasks, to specifically evaluate planning ability on complex long-horizon tasks. We evaluate our method using various LLMs on four datasets in the VirtualHome environment. Our results demonstrate a significant performance enhancement in robotic task planning, showcasing MLDT's effectiveness in overcoming the limitations of existing methods based on open-source LLMs. This research not only advances the field of robotic task planning but also opens avenues for practical applications in complex, real-world scenarios.\footnote{Our code is available at https://github.com/anonym1215/MLDT}

\end{abstract}
\section{Introduction}
The advancement of artificial intelligence (AI) technology in robotics primarily relies on diverse sources of data, including videos and images from cameras, physical parameters from sensors, and the knowledge base in robotics. Therefore, data-driven AI technology has become a popular research topic that aims to enhance the efficiency and adaptivity of AI technology. In recent years, the fusion of data-driven AI technology and robotics has opened new frontiers in task planning, a pivotal technique with broad applications in areas such as home services \cite{DBLP:conf/corl/IchterBCFHHHIIJ22,DBLP:journals/ral/FioriniSPBTC22}, navigation  \cite{DBLP:conf/icra/HuangMZB23}, and manipulation\cite{DBLP:conf/corl/GuhurCPTLS22}. The emergence of large language models (LLMs) has spurred a new era where data-driven insights are leveraged to enhance decision-making and planning capabilities. Specifically, in-context learning \cite{radford2019language,DBLP:conf/nips/BrownMRSKDNSSAA20} retrieves demonstrations from vast datasets and leverages abundant contextual information, such as observations and historical action sequences, to instruct LLMs like ChatGPT\footnote{https://openai.com/chatgpt} in generating task plans grounded in the environment. However, this method is less effective with smaller LLMs \cite{DBLP:journals/corr/abs-2310-12823,DBLP:conf/nips/LiuTMMHBR22} and raises concerns regarding privacy and domain-specific knowledge in closed-source, large-parameter models \cite{wu2023unveiling,ray2023chatgpt}.

%Task planning is a crucial research topic in robotics, with broad applications in areas such as home services \cite{DBLP:conf/corl/IchterBCFHHHIIJ22,DBLP:journals/ral/FioriniSPBTC22}, navigation \cite{DBLP:conf/icra/HuangMZB23}, and manipulation \cite{DBLP:conf/corl/GuhurCPTLS22}. The emergence of large language models (LLMs) has seen their successful application in various tasks \cite{DBLP:journals/www/HuWQMCPA23,DBLP:conf/semweb/TanMLLHCQ23}, including task planning. In-Context Learning (ICL) \cite{radford2019language,DBLP:conf/nips/BrownMRSKDNSSAA20}, which involves constructing prompts with task-related information (environment, action histories) without additional fine-tuning, has been a popular approach with models like ChatGPT\footnote{https://openai.com/chatgpt}. However, this approach is less effective with smaller LLMs \cite{DBLP:journals/corr/abs-2310-12823,DBLP:conf/nips/LiuTMMHBR22} and raises concerns regarding privacy and domain-specific knowledge in closed-source, large-parameter models \cite{wu2023unveiling,ray2023chatgpt}.

\begin{figure}[h]
    \centering
    \includegraphics[width=0.7\linewidth]{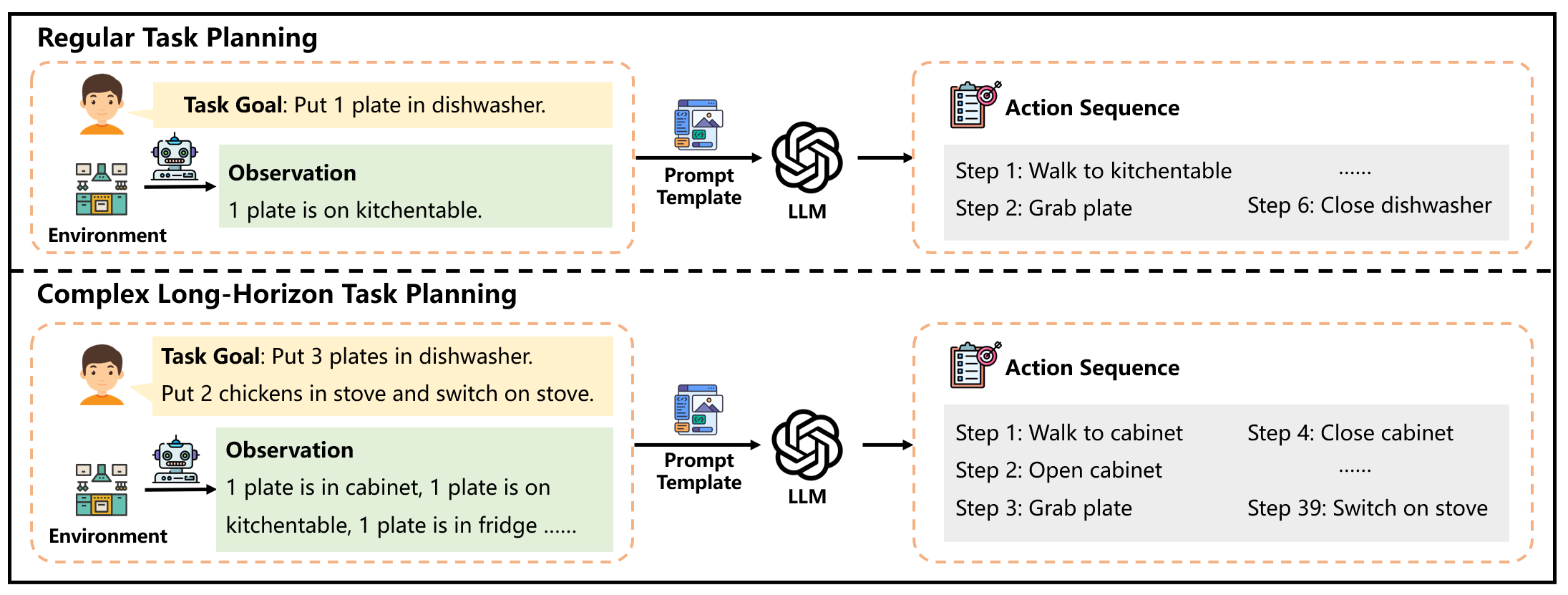}
    \caption{The comparison between regular task planning and complex long-horizon task planning.}
    \label{Figure 1}
\end{figure}

Consequently, alternative efforts focus on fine-tuning small-scale open-source LLMs on task planning data to enhance their planning capabilities \cite{DBLP:journals/corr/abs-2305-10626,DBLP:journals/corr/abs-2307-01848,chalvatzaki2023learning}. They typically organize task-relevant information as input to generate overall task plans. However, several studies \cite{DBLP:journals/corr/abs-2308-03688,DBLP:journals/corr/abs-2307-11088} point out that open-source LLMs possess limited reasoning capacities compared to closed-source LLMs, making them struggle with long-context tasks. Besides, as illustrated in Fig. \ref{Figure 1}, complex long-horizon task planning involves more task goals, more interacting objects, and longer action sequences compared to regular task planning. It requires comprehending substantial context information and generating lengthy action sequences. Therefore, it is a typical long-context task in daily life, such as tidying the room or cooking a meal. Most existing methods incorporate all the task-related information, including task goals and observation, as the input to generate action sequences in a single input-output interaction. Other efforts decompose tasks at the goal-level to mitigate the complexity of the tasks. However, these two strategies are overly difficult for open-source LLMs for they involve excessive context information as inputs and lengthy action sequences as outputs. Previous methods overlook the limited reasoning abilities of open-source LLMs and struggle with this task planning scenario.

In this paper, we focus on complex long-horizon task planning based on open-source LLMs with a small number of parameters (i.e., smaller than 15 billion). The major challenge lies in the contradiction between the limited reasoning capabilities of open-source LLMs and the high complexity of complex long-horizon task planning. Small-scale open-source LLMs exhibit a significant disparity in parameter size compared to closed-source LLMs with over 100 billion parameters. Therefore, they possess limited memory and reasoning abilities \cite{DBLP:journals/corr/abs-2001-08361,DBLP:conf/iclr/WeiBZGYLDDL22}. 

Besides, complex long-horizon task planning involves a high demand for understanding, reasoning, and planning abilities. This phenomenon is similar to humans facing complicated tasks and an excessive number of context information. The intelligence of humans is to break them down recursively until the tasks become sufficiently simple and manageable. To this end, we introduce the \textbf{M}ulti-\textbf{L}evel \textbf{D}ecomposition \textbf{T}ask Planning (MLDT) method, which strategically decomposes tasks at the goal-level, task-level, and action-level. This method effectively simplifies complex long-horizon tasks, making them more manageable for LLMs with limited capacities. Moreover, we enhance the planning abilities of LLMs through instruction tuning and a novel goal-sensitive corpus generation method. This method, powered by ChatGPT, generates a high-quality training corpus by leveraging environmental feedback. We further construct an instruction dataset based on the generated corpus and perform supervised fine-tuning.

%To this end, we propose MLDT, a task planning method that sequentially decomposes tasks at the goal-level, task-level, and action-level. It mitigates the challenges of long-context tasks by gradually transforming complex tasks into simpler ones that LLMs can efficiently address. Besides, we conduct instruction tuning to enhance the planning capability of open-source LLMs. Firstly, to address the lack of training data compatible with our method, we devise a goal-sensitive corpus generation method using ChatGPT. This method aims to generate a high-quality training corpus by incorporating feedback from the environment. Subsequently, we construct an instruction dataset based on the generated corpus and perform end-to-end supervised fine-tuning of open-source LLMs.

To evaluate the performance of our method, we conduct experiments in VirtualHome. Since the complexity of existing dataset is not high enough, we extend the datasets provided by LID \cite{DBLP:conf/nips/LiPPDWF0HAAAM0Z22} and construct a more challenging dataset, LongTasks, to evaluate the planning ability on complex long-horizon tasks. We utilize the original datasets from LID \cite{DBLP:conf/nips/LiPPDWF0HAAAM0Z22} and LongTasks. Experimental results demonstrate that our method achieves the best performance across all datasets and improves over 50\% success rate on the LongTasks dataset.

The contributions of this paper are summarized as follows:

\begin{itemize}

\item We propose MLDT, a novel robotic task planning method. This method decomposes tasks into goal, task, and action levels, addressing the challenge of complex long-horizon task planning. Besides, we conduct instruction tuning to enhance the planning abilities of LLMs. In order to collect high-quality training data compatible with our method, we devise a goal-sensitive corpus generation method based on environmental feedback.

\item We extend the datasets provided by LID and establish a novel dataset, LongTasks, since the complexity of the existing task planning datasets is not high enough. This dataset involves more challenging tasks, incorporating more task goals, more interactive objects, and longer action sequences, to specifically evaluate the planning ability on complex long-horizon tasks.

\item We evaluate our method on four datasets in VirtualHome. Experimental results demonstrate our method effectively enhances the task planning performance of open-source LLMs, particularly for complex long-horizon tasks. 

\end{itemize}

\section{Related Works}

\subsection{Large Language Models for Robotic Task Planning}

The remarkable performance achieved by LLMs across various downstream tasks has spurred extensive research to employ them in robotic task planning. Existing task planning methods based on LLMs can be categorized into two types: (1) In-context learning: Some efforts attempt to construct demonstrations and task-related information into prompts as input, generating task plans grounded in the environment without fine-tuning. Zero-shot Planner \cite{DBLP:conf/icml/HuangAPM22} employs two LLMs, one for generating plans and the other for translating generated plans into executable actions. ProgPrompt \cite{DBLP:conf/icra/SinghBMGXTFTG23} enhances LLMs' capabilities in task planning by adopting a programmatic LLM prompt structure. ReAct \cite{DBLP:conf/iclr/YaoZYDSN023} combines reasoning and acting, allowing LLMs to generate reasoning traces and task-specific actions in an interleaved manner. (2) Fine-tuning: Open-source LLMs with limited parameters struggle to achieve satisfactory results under the in-context learning paradigm. Therefore, some efforts fine-tune LLMs on task planning data to enhance LLMs' planning abilities and incorporate task-related information as input to generate the overall task plans. LID \cite{DBLP:conf/nips/LiPPDWF0HAAAM0Z22} constructs a policy network based on GPT-2 and fine-tunes this network to predict actions interactively. Chalvatzaki et al. \cite{chalvatzaki2023learning} combine a scene graph with task goals as the input for fine-tuned LLMs to obtain plans grounded in the environment. TaPA \cite{DBLP:journals/corr/abs-2307-01848} adopts visual perception models to get observations and constructs a multimodal dataset for grounded tuning LLMs. E2WM \cite{DBLP:journals/corr/abs-2305-10626} fine-tunes LLMs on embodied knowledge collected from world models to improve their embodied knowledge.\\
\indent Our work synergizes these two paradigms, addressing the performance gaps of open-source LLMs through instruction tuning and in-context learning, thereby enhancing their planning and generalization capabilities.

\subsection{Complex Long-Horizon Robotic Task Planning}

Complex long-horizon robotic task planning aims to generate an action sequence to accomplish a specified task goal in robotics. Compared to regular task planning, complex long-horizon task planning involves more task goals and objects, resulting in higher complexity. SayCan \cite{DBLP:conf/corl/IchterBCFHHHIIJ22} employs a pretrained value function to ground the output of LLMs in the environment, enhancing their performances on real-world, abstract and long-horizon tasks. ISR-LLM \cite{DBLP:journals/corr/abs-2308-13724} involves three steps: preprocessing, planning, and iterative self-refinement to improve the feasibility and correctness of the generated plan. Inner Monologue \cite{DBLP:conf/corl/HuangXXCLFZTMCS22} integrates feedback from multiple sources to form an inner monologue, enhancing the performance of LLMs in complex long-horizon tasks. ITP \cite{DBLP:journals/corr/abs-2310-10645} obtains environmental observations based on a visual language model and utilizes a high-level planner to generate a plan, which is then executed by a low-level executor.\\
\indent Unlike previous works incorporating multi-modal information or environmental feedback, we focus on generating executable plans for complex long-horizon tasks based on open-source LLMs under the textual modality without intermediate environment observation. Besides, compared with previous works, the complex long-horizon tasks in this paper involve more task goals, more objects, and longer action steps (i.e., longer than 60), which are much more complicated.

\section{Preliminaries}

Robotic task planning aims to generate an executable plan to achieve task goals within a given environment. A robotic task planning $T$ can be defined as $T=<G, O, A>$, where $G$, $O$, and $A$ represent the task goal, the observation, and all possible actions, respectively. Given the task goal $G$, the observation $O$, and an action set $A = \{a_1, a_2, ..., a_n\}$, the objective of robotic task planning is to generate a $t$-step plan (an action sequence) $\pi = \{a_1, a_2, ..., a_t\}$ to achieve the task goal $G$. For example, as shown in Fig. \ref{Figure 1}, given the task goal $G=$ ``\textit{put 1 plate in dishwasher}'', the observation $O=$ ``\textit{1 plate is on kitchentable}'', and the action set $A=\{walk, grab,... ,close\}$, the task planning model generates an action sequence $\pi=$ ``\textit{walk to kitchentable, grab plate, ..., close dishwasher}'' to achieve the task goal.

%Robotic task planning involves generating an executable plan to achieve task goals within a given environment. A robotic task can be defined as $<S, I, G, T, A>$, where $S$, $I$, $G$, and $A$ represent all possible states, the initial state, the goal state, and all possible actions, respectively. $T$ denotes the transition model, defined as $T: S \times A \rightarrow S$, indicating the impact of executing an action on the environment, resulting in a state transition. Given initial state $I$, goal state $G$, and action set $A = \{a_1, a_2, ..., a_n\}$, the objective of robotic task planning is to generate a $T$ steps plan (an action sequence) $\pi = \{a_1, a_2, ..., a_T\}$ to achieve the transition from the initial state $I$ to the goal state $G$.

To the best of our knowledge, there is currently no strict definition for complex long-horizon task planning. Compared to regular task planning, complex long-horizon task planning involves more task goals and a greater number and variety of objects, resulting in longer action sequences.

\section{Methods}

\subsection{Multi-Level Decomposition Task Planning Method}

\begin{figure}[h]
    \centering
    \includegraphics[width=0.7\linewidth]{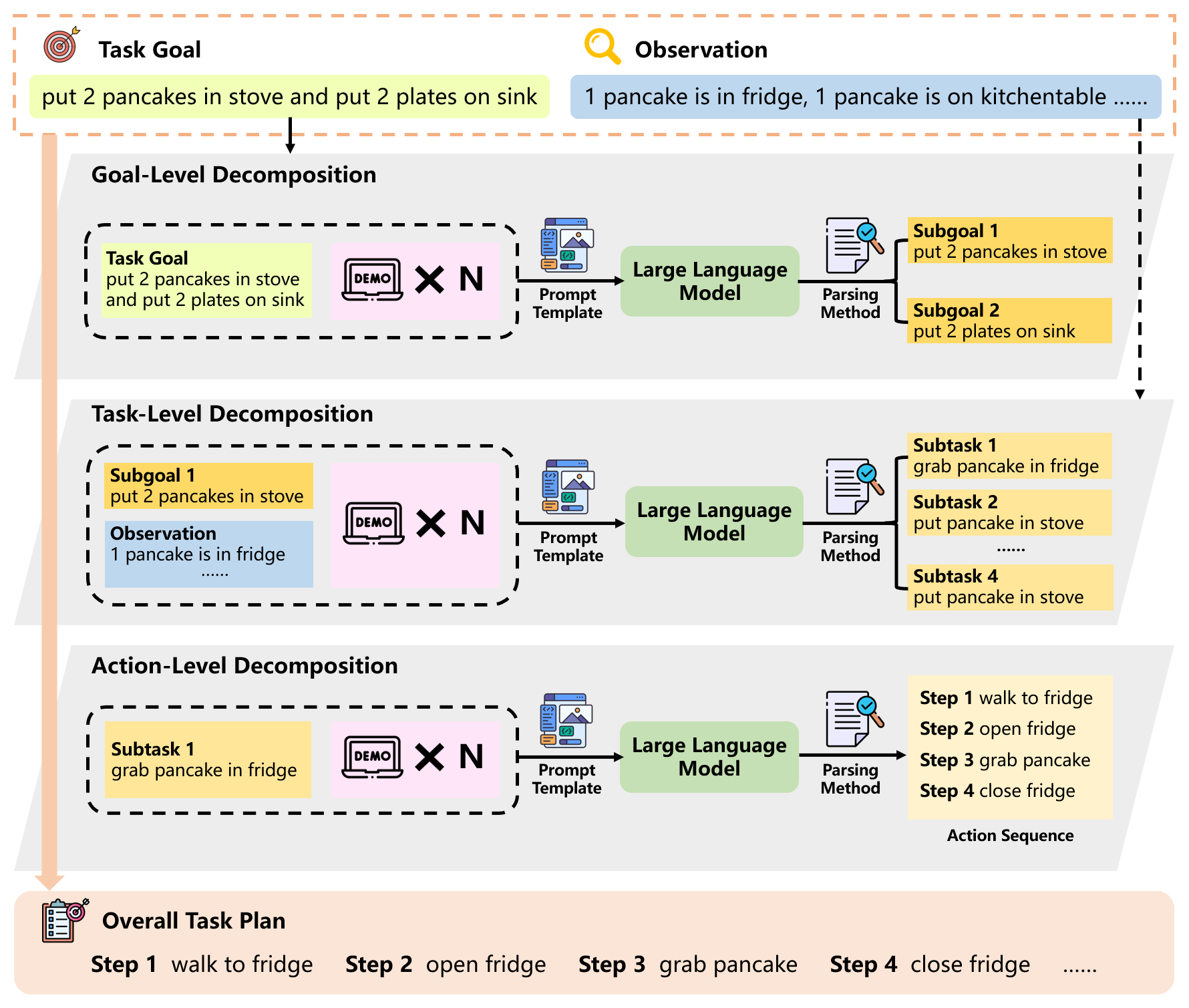}
    \caption{The overview of our proposed method, MLDT.}
    \label{Figure 2}
\end{figure}

To solve the deficiency of open-source LLMs on long-context tasks, we devise MLDT, the multi-level decomposition task planning method. As shown in Fig. \ref{Figure 2}, this method transforms the task planning problem into a three-level decomposition. Compared with the original task planning problem, the complexity of each level decomposition is lower and thus manageable for open-source LLMs. 

\subsubsection{Goal-Level Decomposition} 

Goal-level decomposition aims to break down task goals into several independent subgoals. Given the task goal $G$, we retrieve $N$ examples as demonstrations $demo$ based on the semantic similarity between the examples' task goals and $G$. Each example consists of the input and output of the goal-level decomposition for a given task goal. Then, we employ the goal-level prompt template $P_{goal}$ to transform the demonstrations $demo$ and the task goal $G$ into a prompt, serving as input for LLMs. Finally, We devise a parsing method based on regular expressions to extract $n_{goal}$ individual subgoals $g_1$, $g_2$, ..., $g_{n_{goal}}$ from the output. The goal-level decomposition reduces the number of task goals, thereby decreasing the complexity of task planning.

\begin{equation}
    g_1, g_2, ..., g_{n_{goal}}=LLM(P_{goal}(G, demo))
\end{equation}

\subsubsection{Task-Level Decomposition} 

\begin{figure}[h]
    \centering
    \includegraphics[width=0.7\linewidth]{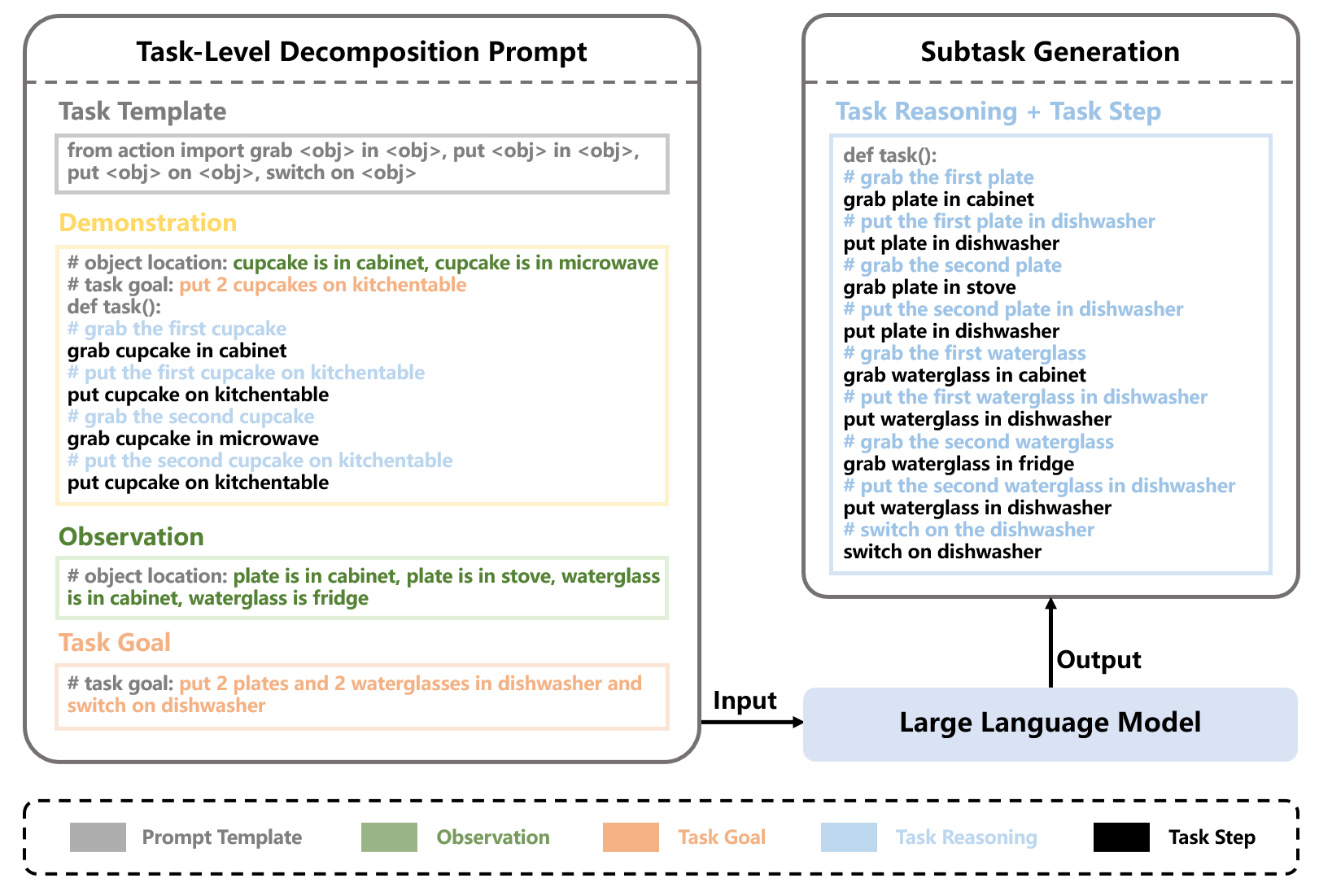}
    \caption{We design a programmatic prompt and generate reasoning traces and actions in an interleaved manner.}
    \label{Figure 3}
\end{figure}

The objective of task-level decomposition is to decompose subgoals into a series of sequential subtasks. We incorporate task templates and observation $obs$ into the input to ensure the generated subtasks are grounded in the environment. Inspired by \cite{DBLP:conf/icra/SinghBMGXTFTG23,DBLP:conf/iclr/YaoZYDSN023}, we design a programmatic prompt and generate reasoning traces and task steps in an interleaved manner, as illustrated in Fig. \ref{Figure 3}. Similar to the goal-level decomposition, given the subgoal $g_c$, we retrieve $N$ examples as the demonstrations $demo$ based on the semantic similarity between the examples' subgoal and $g_c$. The subgoal $g_c$ and the retrieved demonstrations $demo$ are transformed into a prompt based on the task-level prompt template $P_{task}$ as the input. The LLMs' output is parsed through the parsing method based on regular expressions to obtain $n_{task}$ subtasks $t_1$, $t_2$, ..., $t_{n_{task}}$. 

\begin{equation}
    t_1, t_2, ..., t_{n_{task}}=LLM(P_{task}(g_c, obs, demo)), \quad c=1,2, ..., n_{goal}
\end{equation}

\subsubsection{Action-Level Decomposition} 

The action-level decomposition aims to generate an action sequence to complete the given subtask. Similar to the previous two decomposition processes, the subtask $t_{c}$ and the retrieved $N$ demonstrations $demo$ are formulated into a prompt through the action-level prompt template $P_{action}$ as the input. Subsequently, the generated output is processed via the parsing method based on regular expressions to derive the action sequence $a_1$, $a_2$, ..., $a_{n_{action}}$. We aggregate the action sequence of each subtask sequentially to form the overall task plan.

\begin{equation}
    a_1, a_2, ..., a_{n_{action}}=LLM(P_{action}(t_c, demo)), \quad c=1,2, ..., n_{task}
\end{equation}

\subsection{Instruction Tuning for Robotic Task Planning}

\begin{figure}[h]
    \centering
    \includegraphics[width=0.8\linewidth]{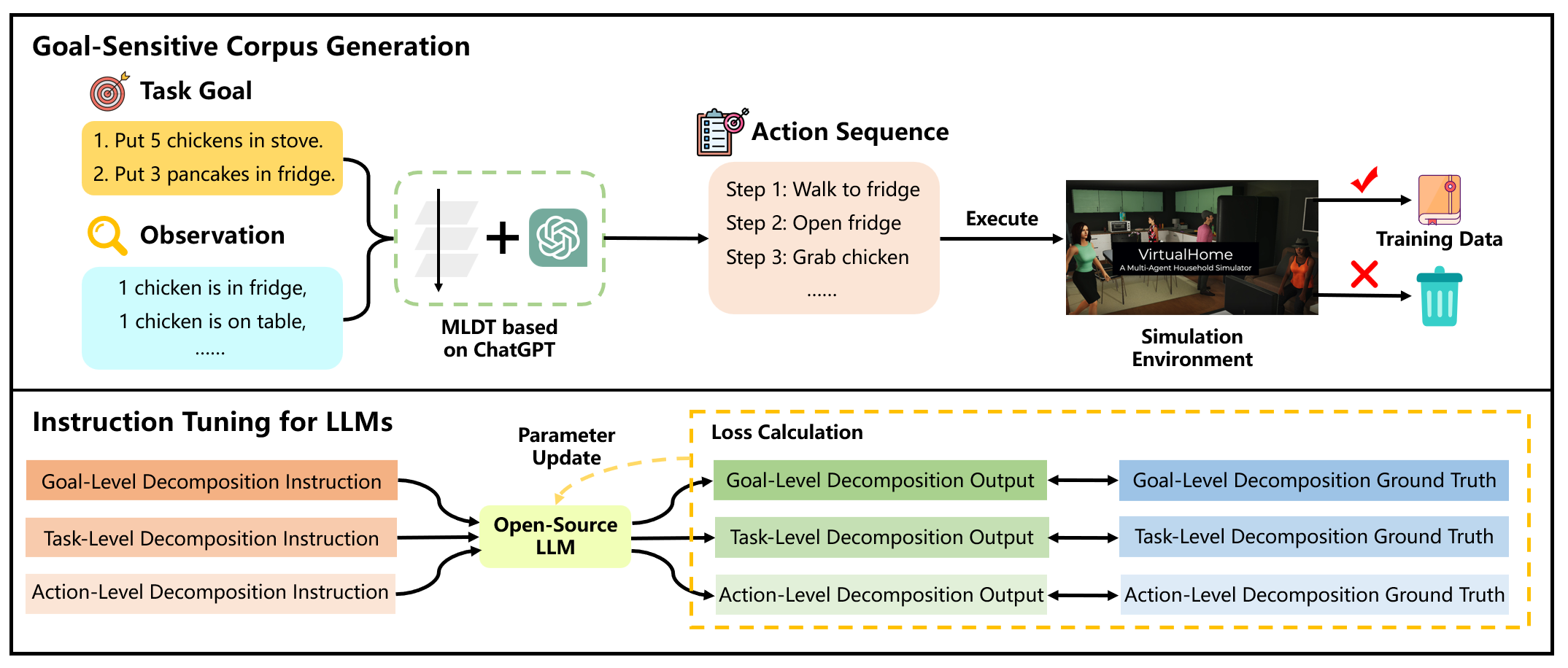}
    \caption{We devise a goal-sensitive corpus generation method to construct a training corpus used for instruction tuning.}
    \label{Figure 4}
\end{figure}

Due to the limited reasoning abilities of open-source LLMs, we conduct instruction tuning to enhance their performance in task planning. As shown in Fig. \ref{Figure 4}, we devise a goal-sensitive corpus generation method to construct a high-quality training corpus which is then used for instruction tuning.

\subsubsection{Goal-sensitive Corpus Generation}

To collect a training corpus for multi-level decomposition, as shown in Fig. \ref{Figure 4}, we devise a goal-sensitive corpus generation method. Inspired by \cite{DBLP:journals/corr/abs-2305-06147,DBLP:journals/corr/abs-2303-15056}, we employ ChatGPT as the primary tool for corpus generation and use it as the foundation model for our proposed method. To optimize the planning capabilities and standardize the output format, we leverage ChatGPT's powerful in-context learning ability to generate task plans by incorporating three demonstrations (i.e., N=3) into the inputs. Due to the absence of annotations, direct evaluation of the generated plan is not feasible. Since task planning aims to achieve task goals, we execute the generated action sequence in a simulation environment like VirtualHome and evaluate the action sequence's quality based on the environmental feedback. If the action sequence is executable and accomplishes the task goal, we consider the plan correct and goal-sensitive and utilize all input-output pairs (excluding demonstrations) as training corpus.

\subsubsection{Instruction Tuning for LLMs}

 The constructed training corpus is mixed thoroughly to ensure variability and richness in the training data. We then frame the input as the instruction and the corresponding output as the ground truth to construct an instruction dataset. Finally, we conduct instruction tuning on this dataset to get LLMs capable of multi-level decomposition. 

\subsection{LongTasks Dataset Construction}

Our work focuses on employing open-source LLMs to address complex long-horizon task planning. However, the current task planning datasets comprise relatively simple tasks that can be accomplished within a few steps. Therefore, the overall complexity is not high enough. To this end, we extend the datasets provided by LID \cite{DBLP:conf/nips/LiPPDWF0HAAAM0Z22} and establish a new dataset called LongTasks. Compared to original datasets, LongTasks includes tasks with more task goals, longer action sequences, and a greater variety and quantity of objects.

\begin{figure}[h]
    \centering
    \includegraphics[width=0.7\linewidth]{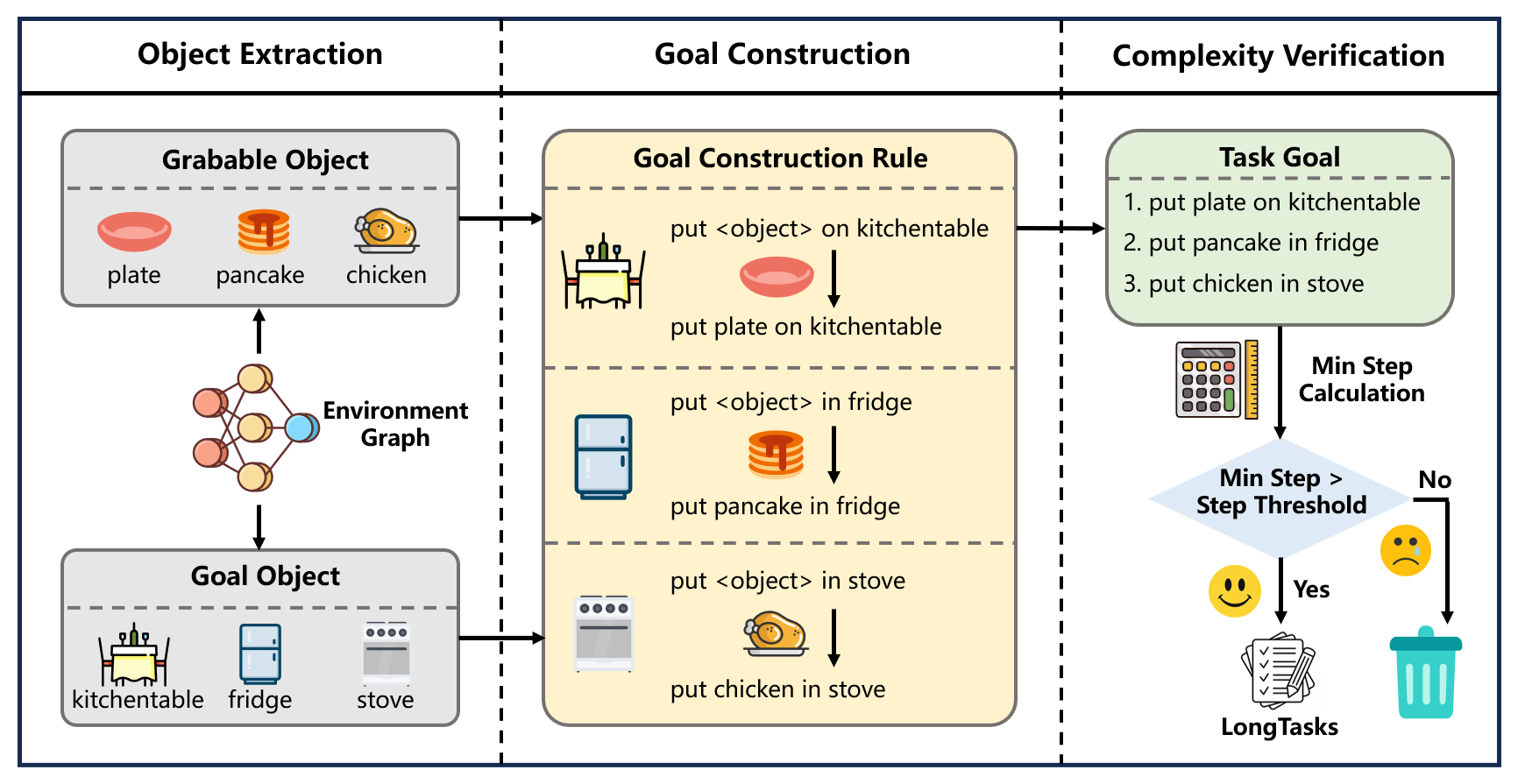}
    \caption{The construction of LongTasks consists of three steps: object extraction, goal construction, and complexity verification.}
    \label{Figure 5}
\end{figure}

The dataset construction process, depicted in Fig. \ref{Figure 5}, unfolds in three steps. First, we extract grabable objects and goal objects from the environment graph. To maintain consistency with the environmental conditions of the original datasets, we only consider objects that appear in them. Then, for each kind of grabable object, we randomly select one goal object and construct a task goal based on the predefined template. Finally, we estimate the minimum action steps based on the task goals, retaining those steps that exceed a predefined step threshold (60 in this work). In the end, we obtain 1,154 samples and conduct a statistical analysis on the original datasets (i.e., In-Distribution, NovelScenes, NovelTasks) and LongTasks. We focus on the average task goals, the average minimum action steps, and the average number and variety of interacting objects. As shown in Table \ref{Table 1}, LongTasks is far more complicated than original datasets.

\begin{table}[h]
    \centering
    \caption{The detailed analysis results of the original datasets and LongTasks.}
\begin{tabular}{l|c|c|c|c}
    \hline
    \textbf{Dataset}& \textbf{Goals Number}&\textbf{Action Step} & \textbf{Objects Number} & \textbf{Objects Variety}\\
    \hline
     \textbf{In-Distribution}& 3.40& 26.58 & 5.32 & 3.57 \\
     \textbf{NovelScenes}& 3.39& 26.27 & 5.32 & 3.56 \\
     \textbf{NovelTasks}& 3.99& 27.02 & 4.97 & 3.40 \\
     \textbf{LongTasks}& \textbf{9.74}& \textbf{77.01} & \textbf{15.79} & \textbf{8.50} \\
    \hline
\end{tabular}
\label{Table 1}
\end{table}

\section{Experiments}

\subsection{Experimental Settings}

\textbf{Environment} We conduct experiments in VirtualHome \cite{DBLP:conf/cvpr/PuigRBLWF018}, a 3D environment that simulates household scenes. This platform comprises rooms and household objects that can be manipulated to execute the generated plan. 

\noindent \textbf{Datasets} Three datasets from LID \cite{DBLP:conf/nips/LiPPDWF0HAAAM0Z22} (i.e., In-Distribution, NovelScenes, and NovelTasks) along with our created dataset, LongTasks, are employed. In-Distribution involves tasks consistent with the training data distribution (not provided by LID \cite{DBLP:conf/nips/LiPPDWF0HAAAM0Z22}), assessing the learning capability. NovelScenes and NovelTasks introduce new scenes and new tasks, evaluating the generalization ability. LongTasks involves complex long-horizon tasks to assess the planning ability for complicated tasks. We randomly sample 100 instances from each dataset for evaluation. We do not retrieve examples as demonstrations for methods using fine-tuned LLMs (i.e., N=0) as the domain information remains consistent between the training and inference phases.

\noindent \textbf{Large Language Models} To assess our method against existing ones, we employed various LLMs of different scales as the backbone. We use Bloom \cite{DBLP:journals/corr/abs-2211-05100} (3B, 7B), ChatGLM \cite{DBLP:conf/acl/DuQLDQY022} (6B) and Llama-2 \cite{DBLP:journals/corr/abs-2307-09288} (7B, 13B). To examine the effectiveness of our method on long-context LLMs, which are trained on the long-context corpus, we incorporate ChatGLM-32k\footnote{https://huggingface.co/THUDM/chatglm3-6b-32k} and LongAlpaca \cite{DBLP:journals/corr/abs-2309-12307}, the long-context versions of ChatGLM and Llama-2, respectively. Additionally, we explore the impact of our method on closed-source LLMs using GPT-3.5\footnote{https://openai.com/chatgpt} and GPT-4\footnote{https://openai.com/gpt-4}.

\noindent \textbf{Training Process} We employ GPT-3.5 as the corpus generator and utilize samples except those used for evaluation in the In-Distribution dataset as the input for the goal-sensitive corpus generation method. To ensure a fair comparison, we construct an instruction dataset for each method using the same corpus generation methods and samples. In order to obtain as much training data as possible, we allow GPT-3.5 to generate plans multiple times until the task is successful and design several rules to correct the generated plans. In the end, the number of training samples for all methods was similar.

\noindent \textbf{Evaluation Metrics} Following \cite{DBLP:conf/icra/SinghBMGXTFTG23,DBLP:conf/icira/ZhangLHZQZLTMSZ23}, we employ \textbf{Success Rate (SR)} and \textbf{Executability (Exe)} to assess whether the plan achieves the task goal and whether the plan is executable. \textbf{SR} is the proportion of tasks that achieve the task goals after executing the plans, and \textbf{Exe} is the proportion of tasks whose plans are executable.

\subsection{Baseline Methods}

We compare MLDT with two categories of methods: 

\textbf{Embodied Planning (Embodied)} constructs task-relevant information into prompts as the input for LLMs fine-tuned on task planning training data, representing the main paradigm in current robotic task planning \cite{DBLP:journals/corr/abs-2305-10626,DBLP:journals/corr/abs-2307-01848,chalvatzaki2023learning}.

\textbf{ReAct} \cite{DBLP:conf/iclr/YaoZYDSN023} generates reasoning traces and actions in an interleaved manner. Although it uses in-context learning in the original work, we set this baseline for we adopt ReAct as the output format in task-level decomposition. 
%Compared to the Embodied baseline, this approach generates reasoning traces for subsequent task goals and rule constraints before action sequences.

To ensure a fair comparison, we use the same corpus generation method for the baselines and our method and fine-tune LLMs on the training data generated by the corresponding method.

\subsection{Research Questions}

In this experiment, we aim to investigate the following research questions:

\textbf{RQ-1} Does our method outperform baselines?

\textbf{RQ-2} How do various methods perform as action steps increase?

\textbf{RQ-3} Is every module of our method necessary?

\textbf{RQ-4} Is our method effective for long-context LLMs?

\textbf{RQ-5} Is our method applicable to closed-source LLMs?

\textbf{RQ-6} Is our method applicable to real-life robots?\\

\noindent \textbf{Does our method outperform baselines? (RQ-1)}

Table \ref{Table 2} shows the overall results of our proposed method and baselines on four datasets. We can observe that: (1) Our method outperforms the baselines by a large margin across all metrics and LLMs, demonstrating its effectiveness. (2) Embodied generally outperforms ReAct. This may be due to the incorporation of the intermediate reasoning process. It increases the context length and results in suboptimal performance for small-scale open-source LLMs with limited reasoning abilities. (3) Llama (13B) exhibits the best performance across different methods, while Bloom performs the worst. Besides the disparity in parameter size, the number of supported languages also affects performance. Consistent with the findings of \cite{DBLP:conf/acl/ConneauKGCWGGOZ20,DBLP:journals/corr/abs-2311-10797}, Bloom supports up to 46 languages, far surpassing Llama (1 language) and ChatGLM (2 languages), leading to its poorer performance on single-language tasks. (4) The disparity between our method and baselines is more significant in LLMs with weaker performance, while the differences among LLMs using our method are relatively smaller. This indicates that our method is more beneficial for LLMs with inferior performance, enhancing the practicality of less powerful LLMs in task planning.

\begin{table}[h]
\centering
\caption{Overall results of our proposed method and baselines on four datasets, where we report SR(\%) and Exe(\%).}
\begin{tabular}{l|cc|cc|cc|cc|cc}
    \hline
    \multirow{2}{*}{\textbf{Method}}& \multicolumn{2}{c|}{\textbf{Bloom$_{(3B)}$}} & \multicolumn{2}{c|}{\textbf{Bloom$_{(7B)}$}}&
    \multicolumn{2}
    {c|}{\textbf{ChatGLM$_{(6B)}$}}&
    \multicolumn{2}
    {c|}{\textbf{Llama$_{(7B)}$}} & \multicolumn{2}{c}{\textbf{Llama$_{(13B)}$}} \\
    \cline{2-11}
     & \textbf{SR} &  \textbf{Exe} &  \textbf{SR} &  \textbf{Exe} & \textbf{SR} & \textbf{Exe} &\textbf{SR} &  \textbf{Exe} &\textbf{SR} & \textbf{Exe}\\
     \cline{1-11}
     ReAct & 54.25& 75.25& 56.25&71.00 & 56.75&81.50 & 69.50&85.25 &71.00&90.25\\
     Embodied & 64.25& 78.00&67.50 &80.00 &61.75 &74.25 &70.00 &81.75 &78.25 & 90.00  \\
     \textbf{MLDT} & \textbf{91.75}& \textbf{93.00}& \textbf{90.75}&\textbf{91.50} & \textbf{92.75}& \textbf{93.50}&\textbf{93.25} & \textbf{93.50}& \textbf{94.25}& \textbf{94.50}  \\
    \hline
\end{tabular}
\label{Table 2}
\end{table}

\begin{figure}[h]
    \centering
    \includegraphics[width=\linewidth]{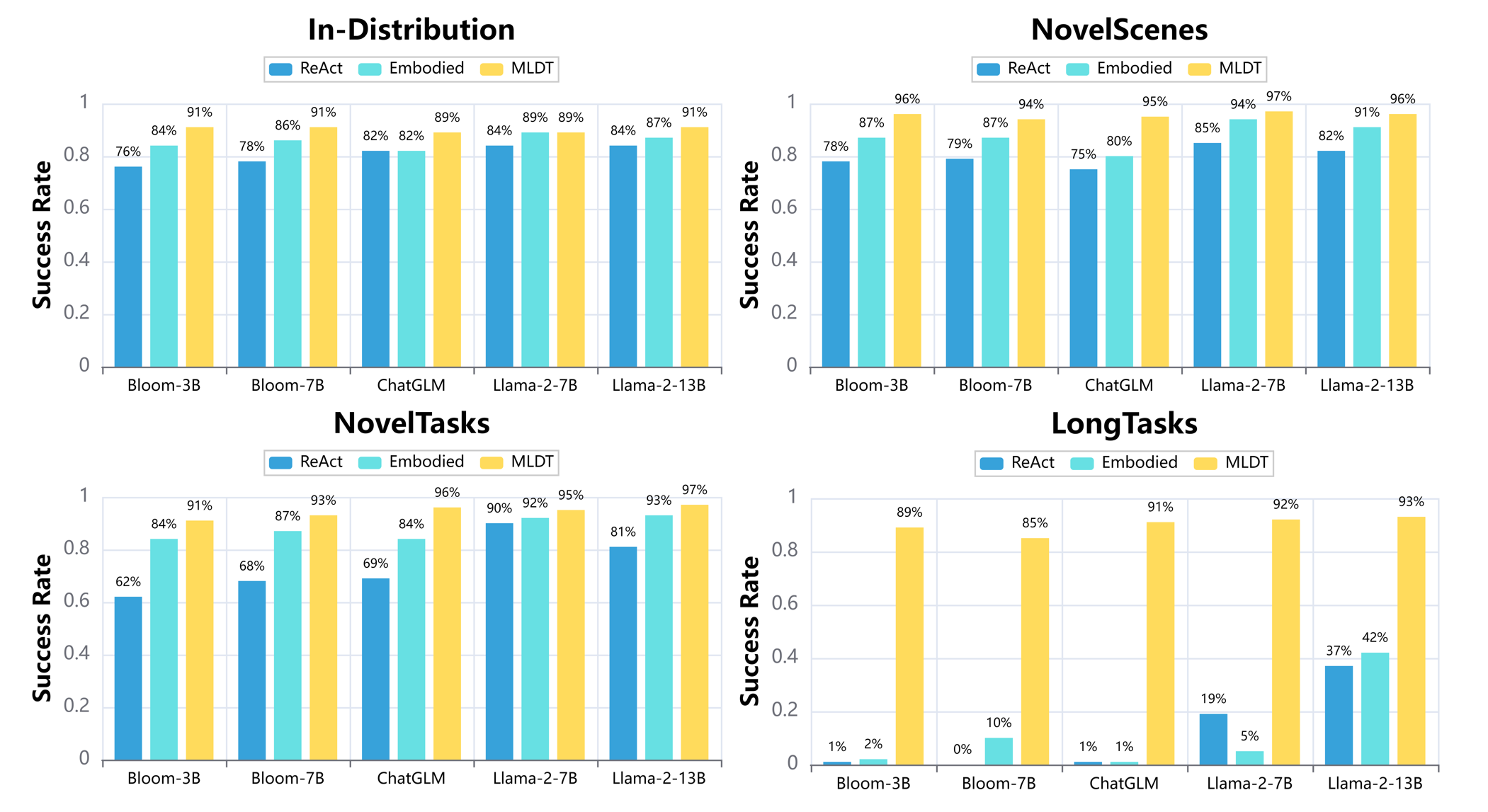}
    \caption{Results of our proposed method and baselines on four datasets, where we report success rate.}
    \label{Figure 6}
\end{figure}

We further analyze the results on each dataset to explore the learning ability, generalization ability, and ability to solve complex long-horizon tasks of different methods. As shown in Fig. \ref{Figure 6}, our method achieves the highest success rate across all datasets, indicating that our method surpasses baselines in various capabilities. It is worth noting that the baselines almost fail on LongTasks, while our method maintains a high success rate. This highlights the effectiveness of our method for solving complex long-horizon tasks.

\noindent \textbf{How do various methods perform as action steps increase? (RQ-2)}

\begin{figure}[h]
    \centering
    \includegraphics[width=0.7\linewidth]{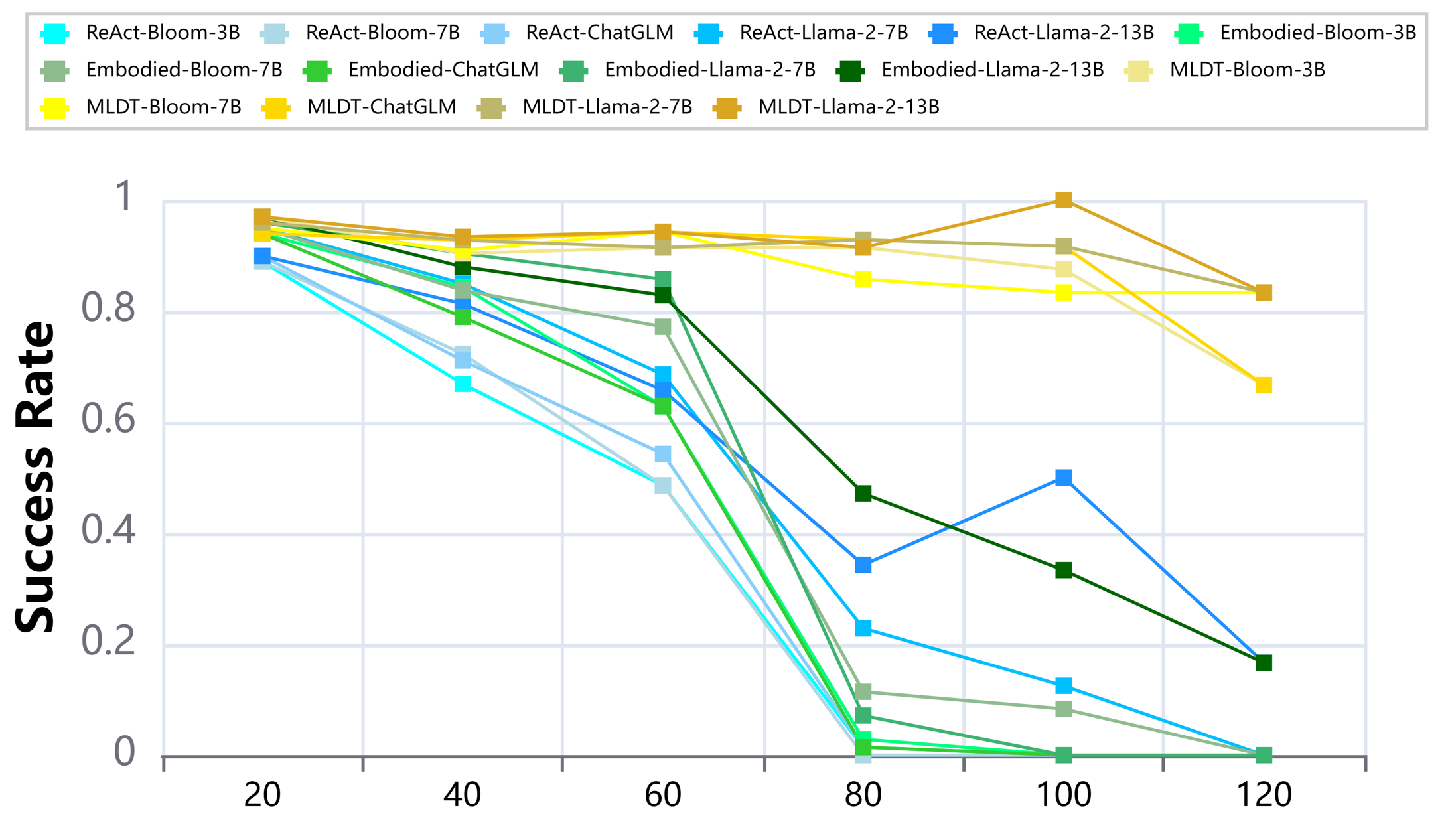}
    \caption{Results of success rate with different action steps. MLDT effectively improves the performance of complex long-horizon tasks.}
    \label{Figure 7}
\end{figure}

To answer this question, we divide the test samples into several groups based on action step length and calculate the success rate of tasks within the corresponding action step interval. As illustrated in Fig. \ref{Figure 7}, our method consistently surpasses the baselines across all action step intervals. The baselines exhibit a significant declining trend as the action step increases. The success rate of task planning generated by the baselines falls below 50\% when the action step is larger than 60. This indicates the ineffectiveness of the baselines on complex long-horizon tasks. In contrast, our method shows fluctuations or a gradual decline in success rate as the action step increases. This suggests that our method is adaptable to tasks of varying complexities, demonstrating strong robustness and generalization capabilities.\\

\noindent \textbf{Is every module of our method necessary? (RQ-3)}

\begin{figure}[h]
    \centering
    \includegraphics[width=\linewidth]{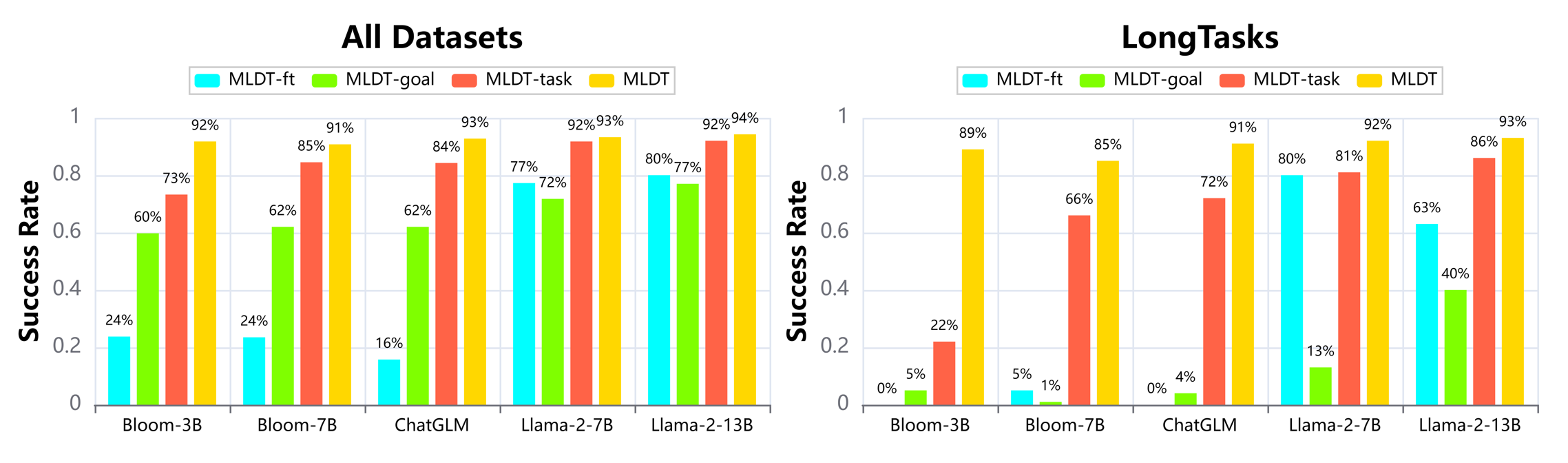}
    \caption{Results of MLDT and its variants on all datasets and LongTasks, where we report success rate.}
    \label{Figure 8}
\end{figure}

We conduct an ablation study to investigate the roles of different modules in our method. Specifically, we design several variants of MLDT: without goal-level decomposition ($\mathrm{MLDT}_{-goal}$), without task-level decomposition ($\mathrm{MLDT}_{-task}$), and without fine-tuning ($\mathrm{MLDT}_{-ft}$). For $\mathrm{MLDT}_{-goal}$ and $\mathrm{MLDT}_{-task}$, we use the same corpus generation method as MLDT and fine-tune LLMs on the generated corpus to ensure a fair comparison. For $\mathrm{MLDT}_{-ft}$, we select three examples (i.e., N=3) to enhance the model's planning abilities and constrain the output format. From the experimental results in Fig. \ref{Figure 8}, we observe that: (1) Relying solely on in-context learning without fine-tuning leads to significant performance degradation, demonstrating the impractical of directly applying in-context learning to small-scale open-source LLMs. Additionally, this indicates the crucial role of fine-tuning in enhancing the planning capabilities of LLMs. (2) Goal-level decomposition has a more significant impact compared to task-level decomposition. We speculate that reducing the goal number in a single task planning instance can significantly lower the complexity of task planning. (3) MLDT outperforms all variants, with a notable performance advantage on LongTasks. This demonstrates that only decomposing tasks at the goal-level or task-level is not enough for open-source LLMs and strongly emphasizes the necessity of each module in our method, especially for complex long-horizon tasks.

\noindent \textbf{Is our method effective for long-context LLMs? (RQ-4)}

\begin{figure}[h]
    \centering
    \includegraphics[width=\linewidth]{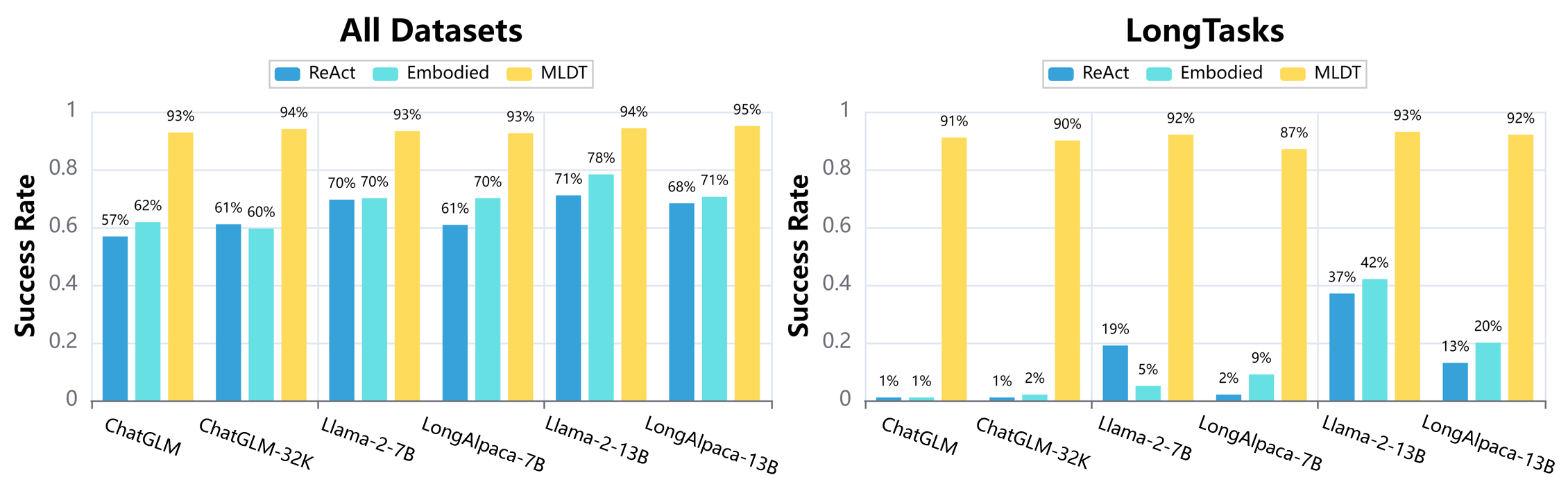}
    \caption{Results of long-context LLMs on all datasets and LongTasks, where we report sucess rate.}
    \label{Figure 9}
\end{figure}

The predefined context size limits LLMs in many long-context tasks. A line of works \cite{DBLP:journals/corr/abs-2307-03170,DBLP:journals/corr/abs-2309-12307} attempts to mitigate this limitation by training LLMs in a longer context. Therefore, we formulate this research question to investigate the performance of long-context LLMs in task planning and evaluate the effectiveness of our method for such LLMs. As illustrated in Fig. \ref{Figure 9}, our method outperforms the baselines by a large margin, particularly on LongTasks. This indicates our method is effective for long-context LLMs. Besides, compared to regular LLMs, long-context LLMs achieve comparable performances in most cases. However, the performance exhibits a noticeable decline when employing the long-context version of Llama in two baseline methods. This suggests that LLMs trained on longer context corpora still fail to address complex long-horizon tasks. We speculate that complex long-horizon tasks necessitate the ability to not only process longer input and output sequences but also grasp longer reasoning chains.

\noindent \textbf{Is our method applicable to closed-source LLMs? (RQ-5)}

\begin{table}[h]
\centering
\caption{Results of our proposed method and baselines using GPT-3.5 and GPT-4, where we report SR(\%) and Exe(\%).}
\begin{tabular}{l|ccccc|ccccc}
    \hline
    \multirow{3}{*}{\textbf{Method}}& \multicolumn{5}{c|}{\textbf{All Datasets}} & \multicolumn{5}{c}{\textbf{LongTasks}}\\
    \cline{2-11}
     & \multicolumn{2}{c}{\textbf{GPT-3.5}} & &\multicolumn{2}{c|}{\textbf{GPT-4}} &\multicolumn{2}{c}{\textbf{GPT-3.5}} && \multicolumn{2}{c}{\textbf{GPT-4}}\\
    & \textbf{SR} & \textbf{Exe} && \textbf{SR} & \textbf{Exe} & \textbf{SR} & \textbf{Exe} && \textbf{SR} & \textbf{Exe}\\
    \hline
    Embodied &38.50&47.00 &&85.50 &87.50 & 0.00&2.00 &&74.00 &80.00\\
    ReAct & 61.25&74.75 && \textbf{90.75}&\textbf{94.00} & 11.00&44.00 &&\textbf{81.00} &\textbf{87.00}\\
    \textbf{MLDT}& \textbf{85.50} & \textbf{88.50}&&32.75&49.25&\textbf{61.00} & \textbf{68.00}&&3.00 &19.00\\
    \hline
\end{tabular}
\label{Table 3}
\end{table}

To investigate this question, we apply MLDT to closed-source LLMs. As fine-tuning closed-source LLMs is not feasible, we select three examples as demonstrations (i.e., N=3). From the experimental results in Table \ref{Table 3}, we can draw the following conclusions: (1) Our method is effective for GPT-3.5 but not applicable to GPT-4. We attribute this to two main reasons. Firstly, GPT-4 is currently one of the most powerful LLMs, achieving superior performance without decomposition. Notably, two baselines based on GPT-4 reach higher success rates than MLDT based on GPT-3.5. Applying our method can lead to error accumulation and performance degradation. Secondly, most errors in GPT-4 are due to not following instructions, but this phenomenon seldom appears in GPT-3.5. This is consistent with the experimental findings in \cite{DBLP:journals/corr/abs-2304-03439,DBLP:journals/corr/abs-2307-10558}, indicating that even LLMs with a huge amount of parameters may have problems not following instructions. (2) ReAct generally outperforms Embodied, which contradicts the results of open-source LLMs. This suggests that ReAct is effective for powerful large-scale LLMs but may not benefit open-source LLMs with smaller scales. We speculate that closed-source LLMs possess strong enough reasoning abilities to handle long-context tasks and benefit from the intermediate reasoning process.\\

\noindent \textbf{Is our method applicable to real-life robots? (RQ-6)}

\begin{figure}[h]
    \centering
    \includegraphics[width=\linewidth]{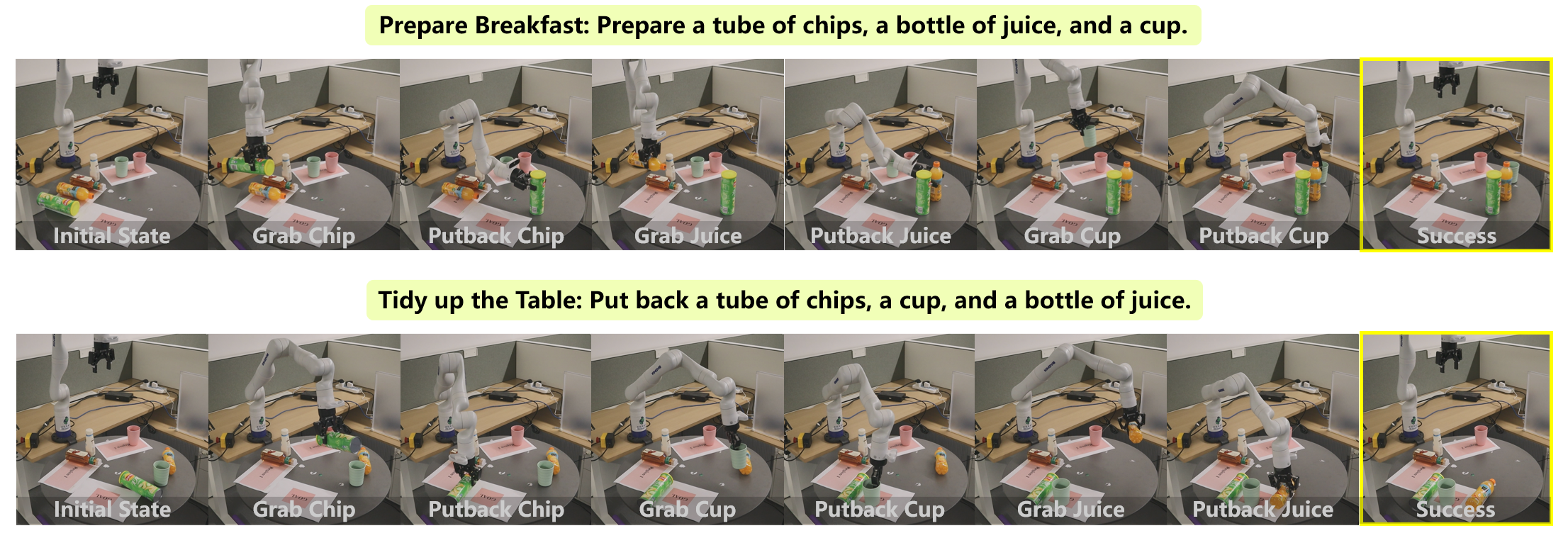}
    \caption{The execution process of two tasks by robots in the real world.}
    \label{Figure 10}
\end{figure}

We deploy MLDT on robots and conduct experiments in the real world. We select two common household scenarios: preparing breakfast and tidying up the table. The fine-tuned Llama (13B) is adopted as the base model due to its best performance. We choose three examples as demonstrations (i.e., N=3) to bridge the gap in domain information between the real world and training data. As shown in Fig. \ref{Figure 10}, the robot successfully achieves task goals. This demonstrates our method is practical in real life and can effectively enhance the capabilities of open-source LLMs in solving everyday tasks.

\section{Conclusion}

In this paper, we propose MLDT, a novel task planning method for small-scale open-source LLMs. This method breaks down tasks at the goal-level, task-level, and action-level to mitigate the challenge faced by open-source LLMs on complex long-horizon tasks. Besides, we devise a goal-sensitive corpus generation method to construct high-quality training data for fine-tuning LLMs, thereby improving their abilities in task planning. Experimental results demonstrate the effectiveness of our method in enhancing the task planning abilities of open-source LLMs. In future work, we plan to apply multi-level decomposition method to a broader range of robotic tasks, with the aim to advance the application of open-source LLMs in the field of robotics.
%This research not only advances the field of robotic task planning but also opens avenues for practical applications in complex, real-world scenarios.

\subsubsection{Acknowledgements.} This work is partially supported by National Nature Science Foundation of China under No. U21A20488. We thank the Big Data Computing Center of Southeast University for providing the facility support on the numerical calculations in this paper.

\bibliographystyle{splncs04}
\bibliography{reference}
\end{document}